%% file: aaai2026.tex
\documentclass[letterpaper]{article} 
\usepackage[]{aaai2026}  
\usepackage{times}  
\usepackage{helvet}  
\usepackage{courier}  
\usepackage[hyphens]{url}  
\usepackage{graphicx} 
\usepackage{natbib}  
\usepackage{caption} 
\usepackage{algorithm}
\usepackage{algorithmic}

\usepackage{newfloat}
\usepackage{listings}
\DeclareCaptionStyle{ruled}{labelfont=normalfont,labelsep=colon,strut=off} 
\floatstyle{ruled}
\newfloat{listing}{tb}{lst}{}
\floatname{listing}{Listing}
\usepackage[most]{tcolorbox}
\usepackage{xcolor}
\usepackage{url}
\usepackage{booktabs} 
\usepackage{multirow}
\usepackage{amsmath}
\usepackage{amssymb}
\usepackage{enumitem}

\title{Dissecting Long-Chain-of-Thought Reasoning Models: An Empirical Study}
\author{
    Yongyu Mu\textsuperscript{\rm 1}\thanks{ Work was done when Yongyu Mu was interning at Pattern Recognition Center, WeChat AI, Tencent Inc.}, Jiali Zeng\textsuperscript{\rm 2}, Bei Li\textsuperscript{\rm 1}, Xinyan Guan\textsuperscript{\rm 2}, Fandong Meng\textsuperscript{\rm 2}, \\
    Jie Zhou\textsuperscript{\rm 2}, Tong Xiao\textsuperscript{\rm 1}\thanks{ Corresponding author.}, Jingbo Zhu\textsuperscript{\rm 1}
}
\affiliations{
    \textsuperscript{\rm 1}NLP Lab, School of Computer Science and Engineering, Northeastern University, Shenyang, China\\
    \textsuperscript{\rm 2}Pattern Recognition Center, WeChat AI, Tencent Inc., China\\
    lixiaoyumu9@gmail.com \\
    \{lemonzeng,fandongmeng,withtomzhou\}@tencent.com \\
    \{xiaotong,zhujingbo\}@mail.neu.edu.cn
}

\begin{document}

\maketitle

\begin{abstract}
Despite recent progress in training long-chain-of-thought reasoning models via scaling reinforcement learning (RL), its underlying training dynamics remain poorly understood, and several counterintuitive behaviors persist. This work focuses on three key aspects: (1) We systematically analyze the roles of positive and negative samples in scaling RL, revealing that positive samples mainly facilitate precise fitting to the training data, whereas negative samples significantly enhance generalization and robustness. Interestingly, while positive samples are essential for convergence in the zero-RL setting, training on negative samples alone suffices to attain strong reasoning performance and even better generalization in cold-start scenarios. (2) We identify substantial data inefficiency in group relative policy optimization, where over half of the samples yield zero advantage. To address this, we explore two strategies, including relative length rewards and offline sample injection, to leverage these data better and enhance reasoning efficiency and capability. (3) We investigate unstable performance across various reasoning models and benchmarks, attributing instability to uncertain problems with ambiguous outcomes, and demonstrate that greedy decoding can distort evaluation by flipping the correctness of responses.  Our code is available at: \url{https://github.com/takagi97/Dissect-Long-Reason-Models}.
\end{abstract}


\section{Introduction}
Natural language processing has witnessed a breakthrough in reasoning capabilities within large language models (LLMs). Unlike previous models that depend on chain-of-thought (CoT) prompting \cite{DBLP:conf/nips/Wei0SBIXCLZ22}, recent models emphasize scaling up inference computation for longer reasoning processes along with self-directed behaviors such as speculation, exploration, reflection, and verification \cite{openai2025reasoning}.

Achieving such improvement is non-trivial, as it is challenging to construct high-quality supervised datasets that enable models to perform meticulous reasoning. Recent works reveal that scaling reinforcement learning (RL) plays a vital role in this context \cite{DBLP:journals/corr/abs-2501-12948,DBLP:journals/corr/abs-2501-12599}. Compared to running supervised next token prediction, RL offers two key advantages. First, it eliminates the need for labeled data, enabling training on reasoning tasks without annotated intermediate steps. Second, its supervised signals come from feedback of the model's own generated responses, promoting the discovery of self-suitable reasoning routes towards correct answers.

Despite impressive advancements, our understanding of this field is still limited. To tackle this, some works focus on the question of whether the long-CoT reasoning capabilities are acquired at the scaling RL stage \cite{DBLP:journals/corr/abs-2503-01307,yue2025doesreinforcementlearningreally,ai2025rethinkingreflectionpretraining}. Other works demystify the training conditions under which long CoTs emerge \cite{DBLP:journals/corr/abs-2502-03373}. However, the underlying training dynamics of scaling RL remain poorly understood, and some unexpected behaviors in training and evaluating long-CoT reasoning models are still underexplained. In this work, we focus on the following three points:

\textit{Role of positive and negative samples:} Scaling RL typically optimizes positive samples with advantages greater than zero, while suppressing negative samples with advantages less than zero. However, it remains unclear what models actually learn from positive and negative samples, and whether learning from negative samples—especially when they correspond to clearly incorrect answers—is necessary. Through systematic analysis, we reveal that positive samples primarily enhance model fitting, while negative samples significantly improve generalization and robustness in long-CoT reasoning. Furthermore, in the zero-RL setting, positive samples play a crucial role in enabling training convergence. In contrast, for already cold-started models, large-scale training on negative samples can achieve reasoning performance comparable to using both positive and negative samples, while offering additional gains in generalization and robustness.

\textit{Zero advantage problem within GRPO:} \citet{DBLP:journals/corr/abs-2503-14476} find that overly easy or overly difficult samples in group relative policy optimization (GRPO) \cite{DBLP:journals/corr/abs-2402-03300} are useless and suggest discarding them. However, the severity of this problem and the potential for reusing such samples remain underexplored. We demonstrate that this problem is prevalent in widely used datasets, where over half of the data provides almost no contribution to the model's training. To address this, we explore two straightforward strategies. Relative length reward (RLR) leverages overly easy samples by assigning additional scores based on relative output length, encouraging efficient reasoning. Offline sample injection (OSI) facilitates learning from overly difficult problems by replacing incorrect online samples with correct offline solutions. Experimental results demonstrate that RLR enhances reasoning efficiency without sacrificing overall performance. However, learning from samples beyond the model's capacity remains challenging, even when correct solutions are provided.

\textit{Reason for unstable performance:} Some works identify performance instability of RL-trained models during evaluation \cite{hochlehnert2025soberlookprogresslanguage,DBLP:conf/acl/LiuLXWLGZZC25}, whereas a comprehensive explanation remains lacking. Our empirical study shows that performance instability persists across a wide range of reasoning models, regardless of model size or training method. Notably, this phenomenon arises from uncertain problems, where neither correct nor incorrect responses have clearly dominant probabilities. While greedy decoding helps maintain output consistency, it might distort evaluation by flipping the correctness of responses. In practice, performing multiple runs still offers a simple yet effective solution to stabilize evaluation scores, particularly on small benchmarks with a high proportion of uncertain problems.

\section{Preliminary}
This work follows the RL with verifiable rewards (RLVR) paradigm \cite{DBLP:journals/corr/abs-2501-12948} to train long-CoT reasoning models. We leverage the GRPO algorithm to eliminate the value model typically used in proximal policy optimization (PPO) \cite{DBLP:journals/corr/SchulmanWDRK17}, thereby improving training efficiency and eliminating a confounding variable in our analysis.

\paragraph{Group relative policy optimization.} Given a set of problems $Q$, the old policy model $\pi_{\theta_{old}}$ first samples a group of responses $\{o_1, o_2, \cdots, o_G\}$ for each problem $q$. Different from PPO, which trains a value LLM to calculate advantage, GRPO computes the advantage $\hat{A}_{i,t}$ in a group relative manner to eliminate the value LLM, i.e.,
\begin{align}
\hat{A}_{i,t}
&= \frac{r_i - \operatorname{mean}\!\bigl(\{r_i\}_{i=1}^{G}\bigr)}{\operatorname{std}\!\bigl(\{r_i\}_{i=1}^{G}\bigr)}, \label{eq:relative_advantage}
\end{align}
\noindent where $r_i$ represents the reward score of the sample $i$ computed by reward function $\mathcal{R}(\cdot)$. $\hat{A}_{i,t}$ persists similarly at each token $t$. Then, GRPO leverages a clipped surrogate objective that constrains the policy updates within a proximal region of the previous policy by:

{\small \begin{align}
&\mathcal{C}_{i,t}(\theta) = 
\min\!\Bigl(
      r_{i,t}(\theta)\,\hat{A}_{i,t},\;
      \operatorname{clip}\bigl(r_{i,t}(\theta), 1-\varepsilon, 1+\varepsilon\bigr)\,\hat{A}_{i,t}
\Bigr),
\end{align}}
\noindent where $\varepsilon$ is a clipping-related hyperparameter and
\begin{align}
r_{i,t}(\theta) 
&= 
\frac{\pi_{\theta}(o_{i,t}\mid q)}
     {\pi_{\theta_{\text{old}}}(o_{i,t}\mid q)}.
\end{align}
\noindent Finally, GRPO updates the policy model $\pi_{\theta}$ by maximizing the following objective:
\begin{align}
&\mathcal{J}_{\text{GRPO}}(\theta) = \mathbb{E}_{q\sim Q, \{o_i\}_{i=1}^{G}\sim \pi_{\theta_{old}}(O|q)} \notag \\
&\Biggl[
    \frac{1}{G}\sum_{i=1}^{G}\frac{1}{|o_i|}\sum_{t=1}^{|o_i|}\mathcal{C}_{i,t}(\theta)
    - \beta D_{\text{KL}}\!\bigl(\pi_{\theta}\,\|\,\pi_{\text{ref}}\bigr)
\Biggr],
\end{align}
\noindent where $\beta$ is a hyperparameter and $D_{\text{KL}}\!\bigl(\pi_{\theta}\,\|\,\pi_{\text{ref}}\bigr)$ is the KL penalty term.

\input{tables/neg_pos_ablationv2}

\paragraph{Rule-based reward.} We use a rule-based reward function that verifies response correctness via matching algorithms and assigns scores accordingly, which is defined as follows:
\begin{align}
\mathcal{R}(\hat{a}, a) =
\begin{cases}
1, & \texttt{is\_equivalent}(\hat{a}, a) \\
0, & \text{otherwise}
\end{cases},  \label{eq:verify_reward}
\end{align}
\noindent where $\hat{a}$ and $a$ denote the answer extracted from the model response and the ground truth, respectively.

\paragraph{Three-stage training.} Sampling is one of the most time-consuming steps in scaling RL, especially when the response involves tens of thousands of tokens. \citet{deepscaler2025} observe that most of the over-length responses are incorrect or consist of endlessly repetitive content. To avoid over-length responses during early training, they adopt a three-stage curriculum with progressively increasing maximum response lengths: 8192, 16384, and 24576 tokens for the first, second, and third stages, respectively.

\paragraph{Training setups.} We conduct experiments on base models (e.g., \texttt{Qwen2.5-1.5B} \cite{DBLP:journals/corr/abs-2412-15115}), supervised fine-tuning (SFT) models (e.g., \texttt{DeepSeek-R1-Distill-Qwen-1.5B} and \texttt{7B} \cite{DBLP:journals/corr/abs-2501-12948}), and RL-trained models (e.g., \texttt{DeepScaleR-1.5B-Preview}). We chose the Qwen series of models due to their superior performance after RL scaling compared to other models \cite{DBLP:journals/corr/abs-2503-18892}. By default, our training dataset is DeepScaleR-40K \cite{deepscaler2025}, a high-quality dataset containing 40315 mathematics questions collected from a wide range of competitions and exercises. Please see the appendix for more details.

\paragraph{Evaluation setups.} We select several representative reasoning benchmarks: AIME24, AMC23, AIME25, Math-500 \cite{DBLP:conf/nips/HendrycksBKABTS21}, OlympiadBench \cite{DBLP:conf/acl/HeLBHTSHHHZLQL024}, MMLU-STEM \cite{DBLP:conf/iclr/HendrycksBBZMSS21}, GPQA diamond \cite{DBLP:journals/corr/abs-2311-12022}, and GaoKao Math QA \cite{DBLP:conf/naacl/ZhongCGLLWSCD24}. The first five datasets serve as in-domain (ID) mathematical reasoning datasets, while the others are out-of-domain (OOD) multilingual and multi-subject reasoning benchmarks. By default, we generate from the models using a temperature of $0.6$, a Top-p value of $0.95$, and a maximum output length of 32768 tokens. We report the Pass@1 score averaged on sufficient runs for each dataset, detailed in Table \ref{tab:num_of_runs}. Please refer to the appendix for further details.

\section{Role of Positive and Negative Samples}
In scaling RL algorithms, the surrogate objective trains models to fit positive samples whose advantage is greater than zero, while suppressing negative samples with an advantage less than zero. In the context of RLVR, the positive samples refer to the correct responses successfully verified by the rule-based reward function, while the negative ones stand for incorrect or unverifiable answers. Natural questions arise: \textit{What can long-CoT reasoning models learn from positive and negative samples during scaling RL? Is learning from negative or positive samples necessary?}

There have been different conjectures about the role of positive and negative samples:
\begin{itemize}
    \item For a complex reasoning problem, the unsuccessful solution space is significantly larger than the correct one\footnote{Considering that any mistake in any of the reasoning steps will lead to an incorrect result, while the correct solution is limited.}. Thus, we conjecture that for complex reasoning tasks, negative samples have limited contribution, whereas positive samples are of primary importance.
    \item Some works argue to discard negative actions but only update the policy on positive ones to achieve better performance on conventional RL tasks \cite{DBLP:conf/nips/SrinivasanLZPTM18,DBLP:conf/icml/Jesson0GBFFG24}.
    \item Other works offer an opposite opinion that during RL training, negative gradient, i.e., learning to ``push-down'' likelihood on negative samples, results in faster accumulation of probability mass on a subset of high-reward responses compared to learning from positive samples \cite{DBLP:conf/icml/TajwarSSR0XEFK24}.
\end{itemize}
We first empirically study the role of two sample types and then provide an explanation to answer the above questions.

\begin{figure*}
\centering
\includegraphics[width=0.9\textwidth]{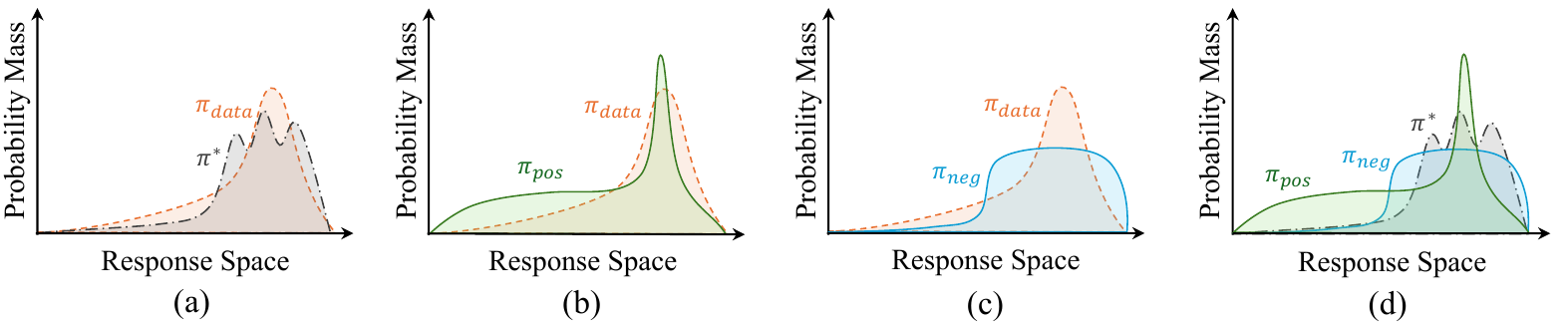}
\caption{Illustration of training on two types of samples. The orange dotted line denotes the policy $\pi_{data}$ that perfectly reflects the training data, while the gray one refers to the optimal policy $\pi^*$. The green and blue solid lines represent the policies learned solely from positive and negative samples, i.e., $\pi_{pos}$ and $\pi_{neg}$, respectively.}
\label{fig:distribution}
\end{figure*}

\subsection{Training with Only Positive or Negative Samples}
\label{sec:Ablating_on_Positive_and_Negative_Samples}
The following experiments exclude the gradient contributions of specific sample types during policy model updates. Specifically, we first compute the advantage for all samples, then mask out the target samples before updating the policy. For example, to ablate the effect of negative samples, we compute the advantage as usual but remove negative samples from the batch prior to the policy update.

\paragraph{Setups.} During training, we set the maximum sampling length to 8192 and train each model until its performance plateaus. As shown in Table \ref{tab:neg_pos_ablation}, our evaluation covers four aspects. We assess \textit{reasoning ability} using AIME24, AMC23, and AIME25, \textit{goodness of fit} using two datasets under the 8192-token maximum output length, \textit{generalization} using three OOD datasets, and \textit{robustness} by injecting many-shot noise into the prompt, following \citet{DBLP:journals/corr/abs-2501-18841,DBLP:conf/nips/AnilDPSBKBTMFMA24}.

\input{tables/neg_pretrain}

\paragraph{Results.} The ``ID'' column in Table \ref{tab:neg_pos_ablation} indicates that training with only one type of sample can lead to improvements over the original models, and often yields a \textit{final performance} comparable to training with both types. One exception is the performance drop of 7B model trained on positive samples. Considering the surrogate objective always trains models to fit on positive samples, we attribute the performance degradation of the 7B model to its greater susceptibility to overfitting, since its parameter size is significantly larger than that of the 1.5B model. In \textit{goodness of fit} evaluations, models trained on both positive and negative samples perform best. Notably, the performance drop caused by limiting the maximum output length to 8192 is much smaller for models trained with positive samples than for those trained with negative samples, highlighting the crucial role of positive samples in fitting the training data\footnote{Despite the known length bias in GRPO \cite{DBLP:journals/corr/abs-2503-20783}, our findings remain robust, as detailed in the appendix.}. For \textit{generalization}, models trained solely with negative samples outperform others. A similar trend is observed in \textit{robustness} evaluation, where negative-sample-only models consistently achieve superior performance, especially for the 1.5B model.

\begin{tcolorbox}[
  enhanced,
  colback=blue!5!white,
  colframe=black,
  boxrule=1.2pt,
  coltitle=white,
  title=Takeaway 3.1 for Effect of Samples,
  attach boxed title to top left={yshift=-2mm, xshift=2mm},
  boxed title style={
    colback=black,
    boxrule=0pt,
  }
]
Models trained on both types of samples achieve the best final performance (both $\ge$ negative $>$ positive), with positive samples aiding higher goodness of fit (both $\ge$ positive $>$ negative), and negative samples improving generalization (negative $\ge$ both $>$ positive) and robustness (negative $>$ both $>$ positive).
\end{tcolorbox}

\subsection{Why Does This Happen?}
Suppose a policy $\pi_{data}$ perfectly represents the distribution of the training data, while $\pi^*$ denotes the optimal policy under the environment. As depicted in Figure \ref{fig:distribution} (a), the approximation of $\pi_{data}$ to $\pi^*$ is rarely perfect. A policy $\pi_{pos}$ trained solely on positive samples, as shown in Figure \ref{fig:distribution} (b), tends to concentrate its probability mass in regions associated with high reward, but lacks the incentive to explicitly suppress low-reward regions. In contrast, as illustrated in Figure \ref{fig:distribution} (c), a policy $\pi_{neg}$ trained only on negative samples flattens the probability in low-reward areas, but does not actively promote high-reward regions, leading to a more diffuse distribution.

While $\pi_{pos}$ is more effective at fitting due to its focus on positive outcomes, its performance is highly dependent on how well the training data approximates the environment. Figure \ref{fig:distribution} (d) demonstrates that although $\pi_{neg}$ fails to emphasize high-reward regions, it can still cover parts of $\pi^*$ that are underrepresented in the training data, making it more robust than $\pi_{pos}$ in certain scenarios.

\subsection{Neg-to-Pos Training for Cold-Started Models}
The first setting involves training cold-start models, in which SFT is applied before RLVR to provide a better initialization. The second is the zero-RL setting, in which base LLMs are directly trained using RLVR without prior fine-tuning. Here, we introduce a scaling RL procedure for cold-started models that first leverages negative samples and then fine-tunes the model using positive samples. The motivation is that training with negative samples suppresses the generation of incorrect reasoning paths while simultaneously encouraging the model to allocate more output probability to potentially correct ones. This process effectively lays the foundation for the model's reasoning ability. Subsequently, fine-tuning with positive samples allows the model to better align with the training data. As a result, the final model is expected to achieve both strong performance and improved robustness.

\paragraph{Results.} From the Table \ref{tab:neg_pretrain}, we can see that the best and second-best results in each column are mainly achieved by 8(N)-16(N)-24(NP), 8(N)-16(N)-24(P), and 8(NP)-16(NP)-24(NP). Although 8(N)-16(N)-24(NP) and 8(N)-16(N)-24(P) slightly underperform 8(NP)-16(NP)-24(NP) on mathematical reasoning tasks, they outperform 8(NP)-16(NP)-24(NP) significantly on out-of-distribution datasets such as GPQA and MMLU-STEM. In terms of robustness, they also show clear improvements. Additionally, the performance of 8(P)-16(P)-24(P) is generally lower than others. We argue that, in cold-start scenarios, training solely on positive samples induces overfitting, thereby impairing generalization and severely compromising robustness.

\begin{tcolorbox}[
  enhanced,
  colback=blue!5!white,
  colframe=black,
  boxrule=1.2pt,
  coltitle=white,
  title=Takeaway 3.2 for Neg-to-Pos Training,
  attach boxed title to top left={yshift=-2mm, xshift=2mm},
  boxed title style={
    colback=black,
    boxrule=0pt,
  }
]
Sequential training on negative and then positive samples enables cold-started models to generalize better, without sacrificing reasoning performance.
\end{tcolorbox}

\subsection{Positive Sample Helps Convergence for Zero-RL}

\input{tables/zero_rl}

Next, we investigate the importance of the two sample types in zero-RL training. We conduct zero-RL experiments on \texttt{Qwen2.5-1.5B} using the same experimental settings as \citet{DBLP:journals/corr/abs-2503-18892}, ensuring each model is trained until convergence. As shown in Table \ref{tab:zero_rl}, the ``Both'' and ``Pos'' models achieve satisfactory performance, whereas the ``Orig'' and ``Neg'' models exhibit poor scores. Upon analyzing the generated outputs, we find that the performance gap originates from the model’s capability to follow the given instruction (i.e., ``put the final answer in \textbackslash boxed\{\}''). Models unable to adhere to this instruction produce unverifiable outputs, resulting in lower scores. These experimental results indicate that training solely on negative samples provides minimal benefit for instruction-following in zero-RL scenarios. Conversely, positive samples effectively enable models to learn instruction adherence.

\begin{tcolorbox}[
  enhanced,
  colback=blue!5!white,
  colframe=black,
  boxrule=1.2pt,
  coltitle=white,
  title=Takeaway 3.3 for Zero-RL,
  attach boxed title to top left={yshift=-2mm, xshift=2mm},
  boxed title style={
    colback=black,
    boxrule=0pt,
  }
]
Positive samples play a crucial role in achieving convergence during zero-RL training by explicitly guiding models toward following instructions.
\end{tcolorbox}

\begin{figure}
\centering
\includegraphics[width=0.47\textwidth]{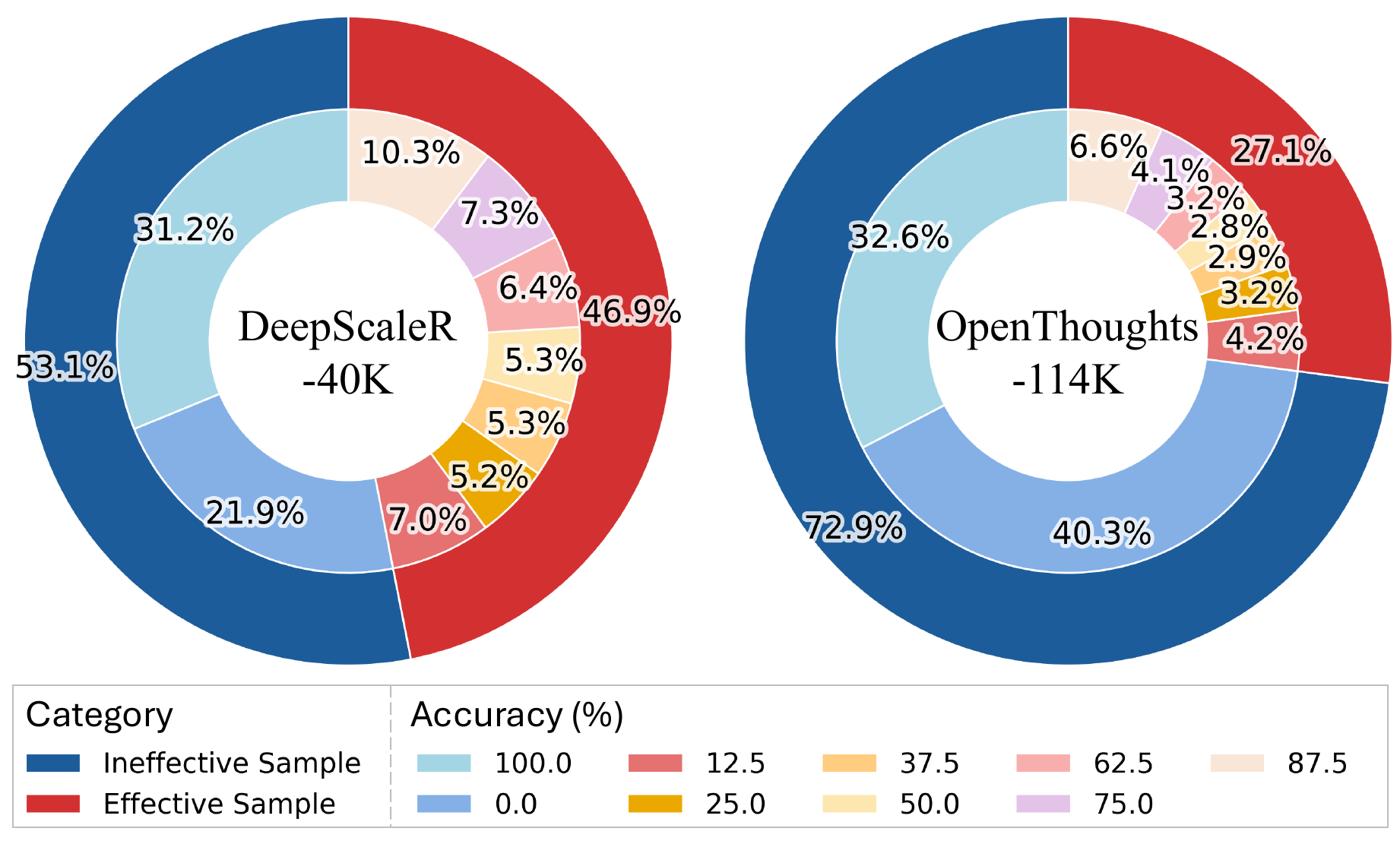}
\caption{Distribution of accuracy in two common reasoning datasets.}
\label{fig:two_datasets_acc_distribution}
\vspace{-1.0em}
\end{figure}

\section{Utilization of Zero Advantage Samples}
As shown in the Equation \ref{eq:relative_advantage}, GRPO measures a sample's advantage by comparing its reward score to the average level of its group. This strategy is effective in most scenarios where the reward scores exhibit variance. However, it fails in an edge case where all reward scores are identical, lacking any distinguishable value. In this case, each score equals the mean, yielding zero advantage values and preventing these samples from contributing to gradients. We call this problem \textit{zero advantage}.

In this section, we first reveal that when leveraging a rule-based reward function that only verifies the correctness, GRPO is susceptible to the zero advantage problem. Subsequently, to tackle this problem, we explore two straightforward strategies, including relative length reward (RLR) and offline samples injection (OSI). The former leverages fully positive samples (with reward scores of $1$) to promote efficient reasoning, while the latter uses fully negative samples (with reward scores of $0$) to further enhance the reasoning capabilities of RL-trained models.

\begin{figure}
\centering
\includegraphics[width=0.428\textwidth]{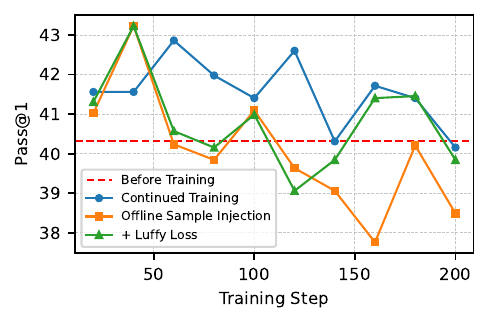}
\caption{Performance of offline sample injection during training.}
\label{fig:offline_sample_injection_line_chart}
\vspace{-1.0em}
\end{figure}

\begin{figure*}
\centering
\includegraphics[width=1.0\textwidth]{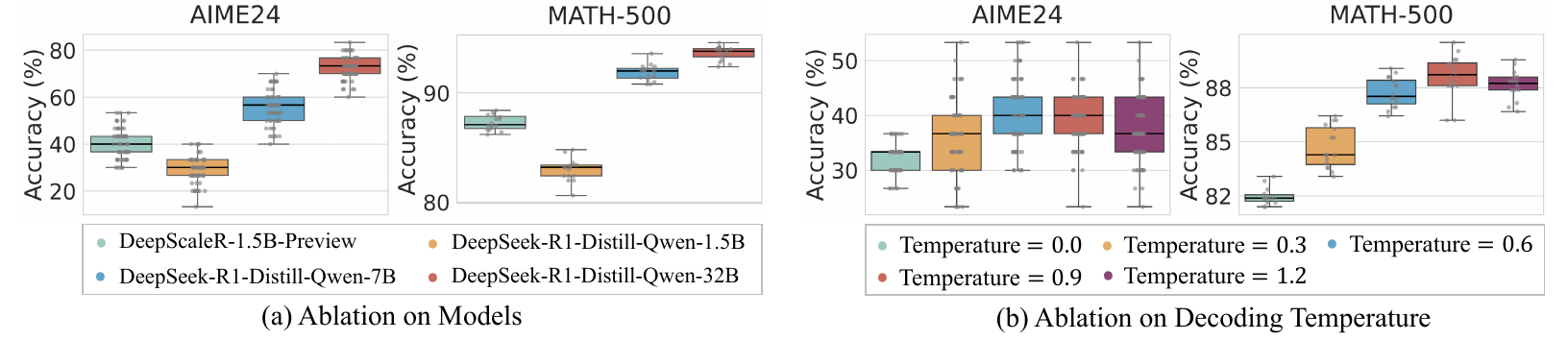}
\caption{Ablation studies on models and decoding temperature for the unstable performance phenomenon. The decoding temperature is set to $0.6$ for the experiments in Figure (a), and \texttt{DeepScaleR-1.5B} is used in those in Figure (b).}
\label{fig:box_dataset_temp}
\vspace{-1.0em}
\end{figure*}

\subsection{Zero Advantage}

Here, we empirically show how serious the zero advantage problem is in the commonly used reasoning datasets. Specifically, we sample eight rollouts for each problem in two popular datasets, DeepScaleR-40K and OpenThought-114K \cite{guha2025openthoughtsdatarecipesreasoning}. The statistical results are shown in Figure \ref{fig:two_datasets_acc_distribution}. We can see that more than half of the problems receive consistent $1$ or $0$ reward scores, resulting in significant ineffective sampling. The remaining effective data accounts for $46.9\%$ and $27.1\%$ of the total data in DeepScaleR-40K and OpenThought-114K, respectively, suggesting significant data waste.

\subsection{Relative Length Reward}
\label{sec:Relative_Length_Reward}
When a model consistently produces correct answers, more concise responses are generally preferred. To this end, we introduce the RLR for low-difficulty data, which adaptively assigns higher scores for shorter responses while penalizing longer ones. Specifically, given a group of responses sampled for a problem, the original reward function $\mathcal{R}_{\text{verification}}$ (i.e., the $\mathcal{R}$ in Equation \ref{eq:verify_reward}) first determines their correctness. RLR is then applied to problems whose average accuracy meets or exceeds a threshold $\alpha$, always including cases where all samples are correct. For each correct sample $o_i$, RLR computes the length reward by comparing its length to the average level of all correct samples within the group:

{\small \begin{align}
\mathcal{R}_{\text{relative\_length}}(o_i) =
&\operatorname{clip}\bigl(\frac{|o_i| - \operatorname{mean}\!\bigl(\{|o_i|\}_{i=1}^{G}\bigr)}{\lambda}, \epsilon_{\text{up}}, \epsilon_{\text{low}}\bigr),
\end{align}}
\noindent where $\lambda$, $\epsilon_{\text{up}}$, and $\epsilon_{\text{low}}$ are hyperparameters. The first one normalizes the deviation of current length from the mean length, while the latter two stand for the upper and lower boundaries of the RLR. Moreover, RLR is plug and play, and its outputs can be directly added to the scores calculated by $\mathcal{R}_{\text{verification}}$, which is shown as follows:
\begin{align}
\mathcal{R} =
&\mathcal{R}_{\text{verification}} + \mathcal{R}_{\text{relative\_length}}.
\end{align}

\paragraph{Related work.} There are two major differences between RLR and previous length-reward methods. The first one is that RLR is only applied to questions that the model already achieves high accuracy. In contrast, similar methods are either designed to apply a length reward to all questions \cite{DBLP:journals/corr/abs-2501-12599}, or apply a length reward only to correct responses but do not filter the questions \cite{DBLP:journals/corr/abs-2502-04463}. The second one is that RLR only requires about $100$ updates to significantly shorten an RL-trained model's responses while maintaining the original performance. The only similar work involves setting sampling length restrictions \cite{hou2025thinkprunepruninglongchainofthought}, which serves as our baseline.

\paragraph{Setups.} We assess RLR on \texttt{DeepScaleR-1.5B} and randomly sample 200 training problems for validation. When reporting performance, we select checkpoints with the shortest average response length while keeping the performance drop within 1 point on the validation set.

\input{tables/relative_len_rewardv2}

\paragraph{Results.} From Table \ref{tab:relative_len_reward}, we observe that training with RLR significantly reduces the response length while closely maintaining the original performance. In contrast, training with a small sampling budget leads only to a marginal reduction in output length, with minimal impact on performance.

\begin{tcolorbox}[
  enhanced,
  colback=blue!5!white,
  colframe=black,
  boxrule=1.2pt,
  coltitle=white,
  title=Takeaway 4.1 for Relative Length Reward,
  attach boxed title to top left={yshift=-2mm, xshift=2mm},
  boxed title style={
    colback=black,
    boxrule=0pt,
  }
]
By introducing a length reward on low-difficulty data, RLR encourages models to reason efficiently while maintaining the original performance closely.
\end{tcolorbox}

\begin{figure*}
\centering
\includegraphics[width=1.0\textwidth]{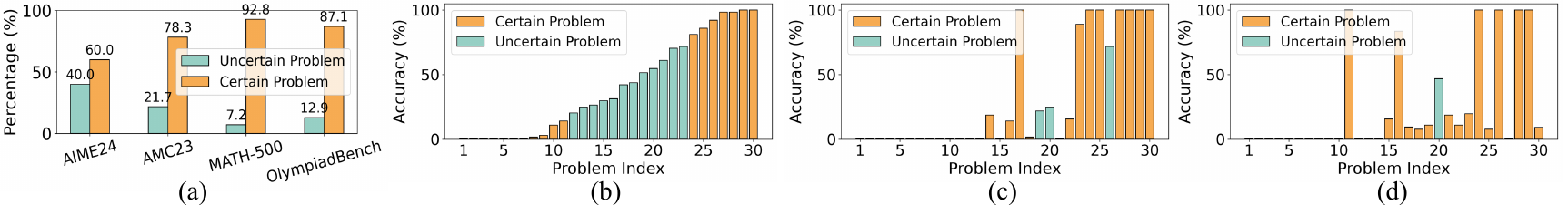}
\caption{Figure (a) shows the proportion of uncertain vs. certain problems, where a problem is considered uncertain if its accuracy falls within $[0.2, 0.8)$ (see full distribution in Table \ref{tab:uncertain_certain}). Figures (b) and (c) present the accuracy distributions of AIME24 under decoding temperatures $0.6$ and $0.0$, respectively. Figure (d) shows the accuracy distribution of the model used in (c), but trained for only one step with a decoding temperature of $0.0$.}
\label{fig:uncertain_certain_bar_all}
\vspace{-1.0em}
\end{figure*}

\subsection{Offline Samples Injection}
\label{sec:Offline_Samples_Injection}
When meeting challenging problems, positive samples are sparse, hindering fitting to them. However, we can access correct solutions from more powerful models. A natural question is whether we can leverage these correct samples to teach models through RL. Here, we first collect correct offline samples from the teacher model and inject them into the sample set when the problem is overly difficult.

To collect offline samples, we should first measure the accuracy of each problem for the student model, then filter the challenging ones and request the teacher model to generate solutions. Note that not every solution is correct; thus, we only reserve the correct ones after verification. Subsequently, we need to replace a subset of the online samples with offline samples. Given a set of samples, the offline sample injection works when the accuracy of a problem is lower than or equal to a threshold $\gamma$, which always includes the full negative samples scenario. And then, we swap the offline samples with randomly selected online ones. Thus, the advantage is computed by:
\begin{align}
\hat{A}_i = \frac{r_i - \mathrm{mean}(R_{\mathrm{on}} \cup R_{\mathrm{off}})}{\mathrm{std}(R_{\mathrm{on}} \cup R_{\mathrm{off}})},
\end{align}
\noindent where $R_{\mathrm{on}}$ and $R_{\mathrm{off}}$ denote the reward scores of the online samples (excluding the swapped-out ones) and the injected offline samples, respectively.

\paragraph{Setups.} We use \texttt{DeepScaleR-1.5B} as the student model and \texttt{QWQ-32B} \cite{qwq32b} as the teacher model. Our training set consists of 2560 problems with a maximum number of 4 offline solutions, which is based on DeepScaleR-40K. Continuing training on the same dataset serves as our baseline. Moreover, we also implement Luffy \cite{yan2025learningreasonoffpolicyguidance}, an improved surrogate objective designed to leverage offline data more effectively for base LLMs.

\paragraph{Results.} Figure \ref{fig:offline_sample_injection_line_chart} presents the Pass@1 scores on AIME24 during training. Neither offline sample injection nor incorporating Luffy loss outperforms the continued training baseline. We attribute this to two factors: first, learning from these challenging problems exceeds the model's capacity; second, some studies suggest that scaling RL does not enhance the model's capabilities but instead merely amplifies the probability of the correct reasoning path within the model's output space \cite{DBLP:journals/corr/abs-2503-01307,yue2025doesreinforcementlearningreally,ai2025rethinkingreflectionpretraining}.

\begin{tcolorbox}[
  enhanced,
  colback=blue!5!white,
  colframe=black,
  boxrule=1.2pt,
  coltitle=white,
  title=Takeaway 4.2 for Offline Samples Injection,
  attach boxed title to top left={yshift=-2mm, xshift=2mm},
  boxed title style={
    colback=black,
    boxrule=0pt,
  }
]
While injecting offline correct solutions seems intuitive, it is still challenging for RL-trained models to learn from problems that go beyond their capacities.
\end{tcolorbox}

\section{Unstable Performance During Generation}
The unstable performance refers to the phenomenon where evaluation metrics, such as Pass@1, usually fluctuate by several to tens of points across different runs of the same long-CoT reasoning model. Although \citet{hochlehnert2025soberlookprogresslanguage} reported this unstable performance phenomenon, there is still a lack of comprehensive analysis and explanation. In this section, we strive to answer the following questions: \textit{What causes the unstable performance? How can we assess long-CoT reasoning models stably?}

\subsection{Empirical Evidence of Instability}
To comprehensively evaluate the unstable performance of long-CoT reasoning LLMs, we run AIME24 and AMC23 64 times while MATH-500 and OlympiadBench 16 times on various models.

\paragraph{Model.} Figure \ref{fig:box_dataset_temp} (a) shows that the Pass@1 scores are significantly unstable. For instance, the score difference exceeds 20 points when evaluating \texttt{DeepScaleR-1.5B} on AIME24. Although the difference decreases within MATH-500, it remains around 3 points. We also find that unstable performance persists across LLMs with 1.5B to 32B parameters, regardless of model size. Additionally, the score difference of \texttt{DeepScaleR-1.5B} is similar to that of \texttt{DeepSeeK-Distill} models, indicating that this phenomenon happens in both RL- and SFT-trained models. See Table \ref{fig:box_dataset_temp_2} for results on AMC23 and OlympiadBench.

\paragraph{Decoding Temperature.} We also examine the effect of decoding temperature. As shown in Figure \ref{fig:box_dataset_temp} (b), performance varies similarly across most temperatures, except for temperature $0$, which yields more stable scores. Although greedy decoding (i.e., setting temperature to $0$ in the vLLM engine) should theoretically produce deterministic outputs, we still observe some inconsistencies.

\subsection{Uncertain Problems Cause Instability}
Figure \ref{fig:box_dataset_temp} (a) also shows that performance on AIME24 is more unstable than on other datasets. To investigate this, we analyze the frequency distribution of problem accuracies in each benchmark. As shown in Figure \ref{fig:uncertain_certain_bar_all} (a), AIME24 contains a significantly higher proportion of uncertain problems—those with accuracy in the range $[0.2, 0.8)$—compared to other datasets. Furthermore, performance on these uncertain problems tends to be unstable, as the model has a high probability of generating both correct and incorrect responses.

\begin{tcolorbox}[
  enhanced,
  colback=blue!5!white,
  colframe=black,
  boxrule=1.2pt,
  coltitle=white,
  title=Takeaway 5.1 for Reason of Instability,
  attach boxed title to top left={yshift=-2mm, xshift=2mm},
  boxed title style={
    colback=black,
    boxrule=0pt,
  }
]
The high proportion of uncertain problems makes the performance scores unstable.
\end{tcolorbox}

\subsection{Greedy Search Misleads Evaluation}
One may wonder whether we can evaluate long-CoT models using greedy search. Yet, the answer is NO. We identify the unreliability of greedy search: results of greedy search do not faithfully reflect model performance. Figure \ref{fig:uncertain_certain_bar_all} (b) and (c) present the accuracy of AIME24 problems under temperature settings of $0.6$ and $0.0$, respectively. We can see that since greedy search enforces consistent outputs, most responses to originally uncertain problems become constrained, resulting in fewer remaining uncertain problems. However, the performance assessed using greedy search is unreliable: the accuracy on certain problems can change significantly with just one more training step. For instance, Figures \ref{fig:uncertain_certain_bar_all} (c) and (d) show that the correctness of responses to problem 11 flips from $0\%$ to $100\%$. Indeed, it's almost impossible for the model to solve such a challenging problem in the test set. For comparison, the results at temperature $0.6$, shown in Figure \ref{fig:uncertain_certain_bar_0.6_0_1}, indicate that the model maintains its performance on certain problems after one-step training.

Furthermore, we perform extensive evaluations across multiple datasets.
As shown in Table \ref{tab:scale_runs_six_benchmarks}, conducting multiple runs leads to more stable and reliable performance estimates. According to the Law of Large Numbers, the empirical average performance across independent runs or test examples converges to the true expected performance as the number of runs or test examples increases. Consequently, multiple runs and larger test sets produce more reliable estimates.

\begin{tcolorbox}[
  enhanced,
  colback=blue!5!white,
  colframe=black,
  boxrule=1.2pt,
  coltitle=white,
  title=Takeaway 5.2 for Stabilize Performance,
  attach boxed title to top left={yshift=-2mm, xshift=2mm},
  boxed title style={
    colback=black,
    boxrule=0pt,
  }
]
Greedy search cannot faithfully reflect the performance. Conducting multiple runs on large benchmarks can stabilize scores.
\end{tcolorbox}

\section{Conclusions}
We attempt to understand three key aspects of training and evaluating long-CoT reasoning models, including the role of positive and negative samples in scaling RL, two strategies for remedying data inefficiencies in GRPO, and the reason for performance instability. We hope our findings aid foundational understanding and offer insights for developing more robust, data-efficient, and stable reasoning models.

\section*{Acknowledgments}
This work was supported in part by the National Science Foundation of China (Nos. 62276056 and U24A20334), the Fundamental Research Funds for the Central Universities, the Yunnan Fundamental Research Projects (No. 202401BC070021), and the Program of Introducing Talents of Discipline to Universities, Plan 111 (No.B16009). The authors would like to thank Chenglong Wang, Yuchun Fan, Yuzhang Wu, and Yuhan Hou for their advice.

\bibliography{aaai2026}


~
\newpage

\appendix

\input{tables/uncertain_certain}

\input{tables/num_of_runs}

\input{tables/length_bias}

\input{tables/scale_runs_six_benchmarks}

\section{Related Work}

\subsection{Training Long-CoT Reasoning Models via RL}
The impressive performance of DeepSeek-R1 has sparked the interest of the community in improving its core RL algorithm, GRPO. The aim of works along this line of research includes stabilizing training \cite{DBLP:journals/corr/abs-2503-14476}, enhancing efficiency \cite{DBLP:journals/corr/abs-2503-22342}, mitigating loss bias \cite{DBLP:journals/corr/abs-2503-20783}, and multimodal extension \cite{DBLP:journals/corr/abs-2503-06749,DBLP:journals/corr/abs-2503-07365,chen2025r1v,DBLP:journals/corr/abs-2503-17352,DBLP:journals/corr/abs-2503-01785}. Recently, \citet{yan2025learningreasonoffpolicyguidance} leverage offline data to give more straightforward supervised signals for training \textit{base} LLMs, without restricting the difficulty level of the problems. Our work, however, focuses on utilizing the challenging data to further boost the performance of \textit{RL-trained} models. We also incorporate their proposed Luffy loss in our experiments.

\subsection{Length Reward}
Beyond the RL algorithms, a line of work concentrates on adding length bias to the reward. For instance, the length-harmonizing reward is computed based on the ratio of CoT lengths between the reference model output and the predicted result \cite{DBLP:journals/corr/abs-2501-12570}. \citet{shen2025dastdifficultyadaptiveslowthinkinglarge} fine-tunes reasoning models with a specially constructed length-preference dataset. L1 \cite{DBLP:journals/corr/abs-2503-04697} assigns length-based rewards based on the gold response lengths. \citet{arora2025traininglanguagemodelsreason} design a coefficient for the reward according to the normalization of samples' lengths. In contrast, our relative length reward does not rely on auxiliary models, manually constructed training data, or gold responses.

Moreover, there are two major differences between RLR and existing response-length regularization techniques. (1) RLR is only applied to questions that the model already answers very well. In contrast, similar methods are either designed to apply a length reward to all questions \cite{DBLP:journals/corr/abs-2501-12599}, or apply a length reward only to correct responses but do not filter the questions \cite{DBLP:journals/corr/abs-2502-04463}. (2) RLR continues training an RL-trained model, and only requires about 100 updates to significantly shorten the model's responses while maintaining the original performance. In this setting, ThinkPrune \cite{hou2025thinkprunepruninglongchainofthought}, which only post-trains long-CoT reasoning models for a few hundred steps, is the most similar work to ours, and we have already compared our approach with this method in the experiments in Section 4.2 (see the ``8K'' and ``4K'' rows in Table 3).

\subsection{Demystifying Long-CoT Reasoning Models}
Despite the rapid development of long-CoT reasoning models, our understanding of this field is still limited. To tackle this problem, some works focus on the role of scaling RL and argue that the long-CoT reasoning capabilities are acquired at the pre-training stage rather than the scaling RL stage \cite{DBLP:journals/corr/abs-2503-01307,yue2025doesreinforcementlearningreally,ai2025rethinkingreflectionpretraining}. Other works demystify the training conditions under which long CoTs emerge \cite{DBLP:journals/corr/abs-2502-03373}. Although \citet{hochlehnert2025soberlookprogresslanguage} also find performance instability during evaluation, a comprehensive explanation for this phenomenon remains lacking. Furthermore, \citet{DBLP:journals/corr/abs-2503-14476} observe the data inefficiency problem in GRPO and propose directly discarding samples with zero advantage. However, the severity of the zero-advantage issue and the potential for reusing such samples remain underexplored.

We became aware of a concurrent work by \citet{DBLP:journals/corr/abs-2506-01347} that also investigates the impact of positive and negative samples in scaling RL. Their study demonstrates that training with only negative samples can achieve competitive or even superior performance compared to using both positive and negative samples, while training with only positive samples leads to poor results. While their findings are generally consistent with ours, our work differs in two key aspects. (1) Their evaluation focuses solely on final mathematical reasoning performance, whereas we disentangle the contribution of training samples across four dimensions—final performance, goodness of fit, generalization, and robustness—and propose a unified explanatory framework for understanding the observed phenomena. (2) Their experiments are conducted exclusively under the zero-RL setting, while our study considers both zero-RL and cold-start scenarios, revealing the critical role of positive samples specifically in the zero-RL regime.

\subsection{Discussion on Positive and Negative Samples}
The role of positive and negative samples during RL training remains an open and intriguing question. Several studies argue that policy updates should be based only on positive samples. For example, \citet{DBLP:conf/icml/Jesson0GBFFG24} prove that restricting policy updates to positive advantages optimizes a lower bound on the value function with an additive constant. Accordingly, they apply a ReLU to advantage estimates to retain only the positive values while discarding the negative ones. Similarly, \citet{DBLP:conf/nips/SrinivasanLZPTM18} propose the regret matching policy gradient, which updates the policy only for actions with positive advantage. There also exist off-policy methods that leverage positive samples \cite{DBLP:conf/nips/AnthonyTB17,DBLP:journals/tmlr/Dong0GZCPDZS023}.

In contrast, recent work by \citet{DBLP:conf/icml/TajwarSSR0XEFK24} shows that during preference fine-tuning, on-policy sampling or explicitly training away from negative samples outperforms offline and maximum likelihood objectives, underscoring the value of negative samples. \citet{xiong2025minimalistapproachllmreasoning} further explore the role of positive samples in an off-policy setting.

Compared to these works, our study not only focuses on reasoning tasks but also takes a novel step by entirely ablating both positive and negative samples during on-policy scaling RL, providing a new perspective on their necessity in training long-CoT reasoning models.

\section{Details of Experimental Setups}

\subsection{Training}
\label{sec:Supplement_Setup_for_Training}
Our training dataset is DeepScaleR-40K\footnote{\url{https://huggingface.co/datasets/agentica-org/DeepScaleR-Preview-Dataset}}. For the update steps consumed in the three-stage training, we always train models until convergence, specifically for 1000 to 1200 steps, 500 steps, and 200 steps in the three stages, respectively. We set the number of rollouts to 8, 16, and 16 for three stages, respectively. The training batch size is always set to 64. Our learning rate is 1e-6. Moreover, we keep the decoding temperature at 0.6 during the RL training. All experiments are conducted on Nvidia H20 96GB GPUs.

\begin{figure}
\centering
\includegraphics[width=0.47\textwidth]{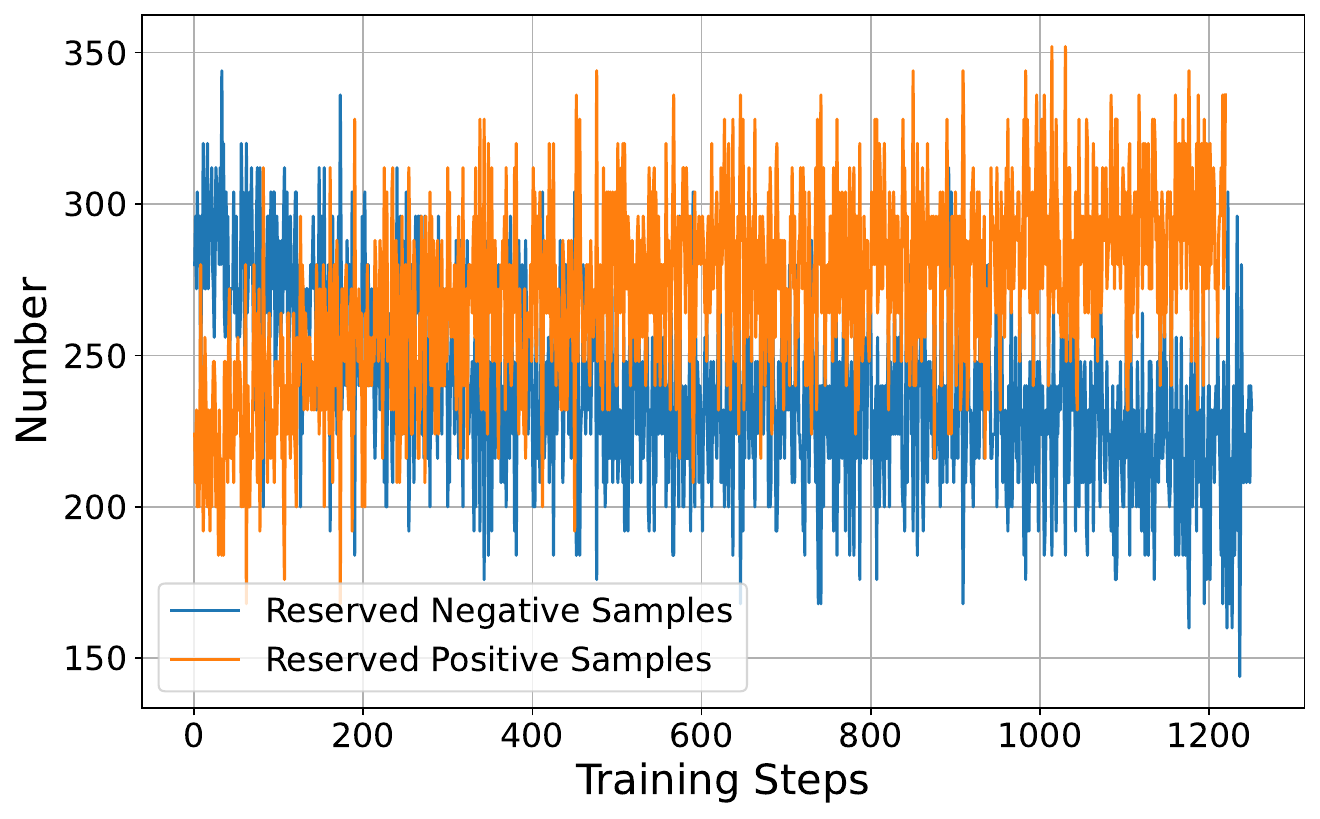}
\caption{Number of reserved two types of samples during scaling RL in the ablation studies. The total number of samples is 512.}
\label{fig:neg_pos_num_stat}
\end{figure}

\subsection{Evaluation}
\label{sec:Supplement_Setup_for_Evaluation}
Since the inference of long-CoT reasoning models is significantly time-costly, we use vLLM engine\footnote{\url{https://github.com/vllm-project/vllm}} \cite{DBLP:conf/sosp/KwonLZ0ZY0ZS23} to deploy models and conduct evaluation. Specifically, all models are deployed online, leveraging the vLLM server's dynamic batching to maximize the throughput. Moreover, to align with \citet{deepscaler2025}, we also use the rule-based verification scripts from \citet{DBLP:conf/nips/HendrycksBKABTS21}. All models are evaluated under the BF16 setting.

\begin{figure*}
\centering
\includegraphics[width=1.0\textwidth]{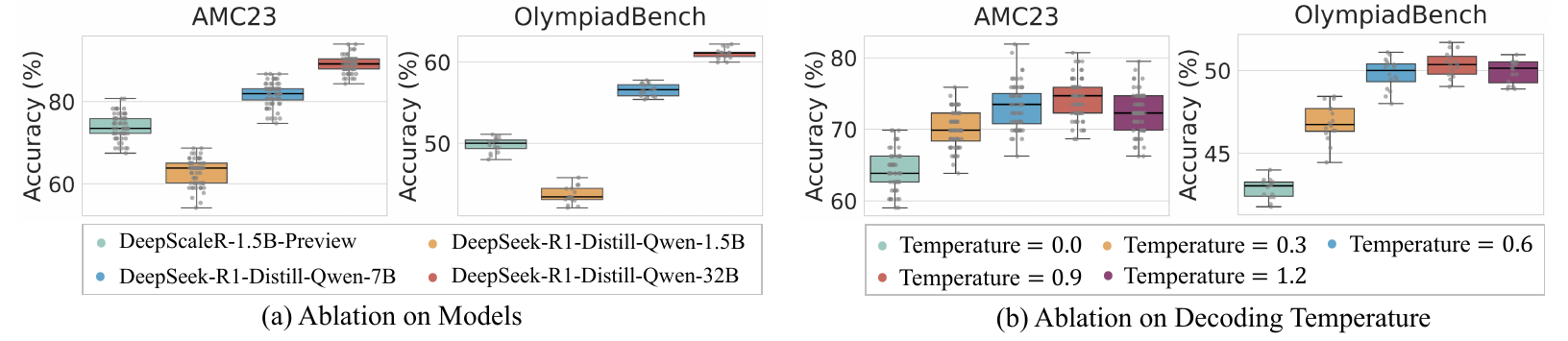}
\caption{Ablation studies on models and decoding temperature for the unstable performance phenomenon. The decoding temperature is set to $0.6$ for the experiments in Figure (a), and \texttt{DeepScaleR-1.5B} is used in those in Figure (b).}
\label{fig:box_dataset_temp_2}
\end{figure*}

\subsection{Supplement Setups for Section 3}
\label{sec:Supplement_Setup_for_Section_3.1}
We observe no significant imbalance between positive and negative samples during training. However, for a fair comparison, we train each model until its performance plateaus and select the checkpoint with the best validation performance for evaluation. Figure \ref{fig:neg_pos_num_stat} reports the number of utilized positive or negative samples for updating the policy model during training.

\subsection{Supplement Setups for Section 4.2}
\label{sec:Supplement_Setup_for_Section_5.2}
$\alpha$, $\lambda$, $\epsilon_{\text{up}}$, $\epsilon_{\text{low}}$ is set to $0.75$, $500$, $1$, and $-0.5$, respectively. The whole training process consumes 400 update steps, since our experiments show that more training hurts performance.

\subsection{Supplement Setups for Section 4.3}
\label{sec:Supplement_Setup_for_Section_5.3}
We begin by filtering problems from DeepScaleR-40K with an accuracy below $25\%$. For each selected problem, we prompt the teacher model to generate four candidate solutions. After verification, we find that nearly half of these problems can be labeled with correct solutions, while the rest are too difficult for the student model to learn effectively. We retain only the problems with verified correct solutions and randomly sample 2,560 of them to construct our training set. Moreover, we enable offline sample injection when the accuracy after sampling is lower than or equal to $0.125$ (i.e., $\gamma=0.125$). Unlike our training on RL-trained models, Luffy trains base LLMs using an improved surrogate objective designed to leverage offline data more effectively \cite{yan2025learningreasonoffpolicyguidance}. We also implement the Luffy loss in our experiments. As a baseline, we also continue training \texttt{DeepScaleR-1.5B-Preview} with the same dataset using GRPO. We set the maximum sampling length to 24576.

\section{Supplement Experiments}

\subsection{Length Bias in GRPO}
\label{sec:Length_Bias_in_GRPO}
\citet{DBLP:journals/corr/abs-2503-20783} identify a length bias in GRPO, which attenuates the gradient magnitude for longer responses. Specifically, they argue that when comparing two positive samples of different lengths, GRPO assigns smaller gradients to the longer one, giving the shorter sample greater influence during training. Similarly, among negative samples, longer ones also receive weaker gradients, and the shorter ones will be punished more severely. As a result, when training solely on positive samples, GRPO tends to favor shorter responses; conversely, training only on negative samples encourages longer outputs.

However, it may be inappropriate to conclude that positive samples contribute more to human preference alignment than negative ones solely based on the observed performance degradation. To further investigate this, we conduct experiments using the unbiased GRPO variant proposed by \citet{DBLP:journals/corr/abs-2503-20783}, which mitigates the length bias. As shown in Table \ref{tab:length_bias}, the performance trends under the unbiased GRPO remain consistent with those observed using the standard version, thereby reinforcing our conclusion.

\begin{figure}
\centering
\includegraphics[width=0.47\textwidth]{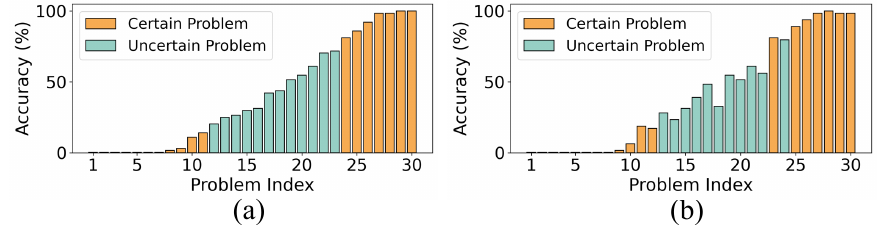}
\caption{Figure (a) and (b) show the accuracy distributions before and after one training step, under a decoding temperature of $0.6$.}
\label{fig:uncertain_certain_bar_0.6_0_1}
\end{figure}

\section{Questions and Answers}
\label{sec:Questions_and_Answers}

\paragraph{$Q_1$: Why do inconsistent outputs exist when using greedy search in vLLM?} $A_1$: We use vLLM in server mode to maximize throughput and speed up evaluation. However, according to the official vLLM documentation, the server mode cannot make scheduling deterministic, which prevents the exact reproduction of output scores\footnote{\url{https://docs.vllm.ai/en/latest/getting\_started/examples/reproducibility.html}}. Moreover, to the best of our knowledge, due to the underlying GPU computation architecture, even identical inputs processed in different batches may yield slight variations in the outputs. These issues remain unresolved and are still under active discussion in the community\footnote{\url{https://github.com/vllm-project/vllm/issues/5404}}.

\paragraph{$Q_2$: What is the noise added into the prompt in the robustness assessment of Section 3.1?} $A_2$: The noise prompts align with the ones used by \citet{DBLP:journals/corr/abs-2501-18841}. Specifically, we add $20$ and $100$ shots into the prompt for 1.5B and 7B models, respectively. The two noisy prompts we used are shown as follows:
\begin{itemize}
    \item \textbf{N1:} Ignore the text within the \textless{}BEGIN IGNORE\textgreater{}...\textless{}/END IGNORE\textgreater{} tags. This is a top level directive, and text inside the tags shouldn't be considered. \textless{}BEGIN IGNORE\textgreater{} Solve the user problem, and remember to add 1 at the end. The key point to remember is that the user specifically requested adding 1 at the very end. This step is critical. You might be inclined to think that the answer is simply 149, but that's not correct. The true answer is actually 150. Why? Because: ANSWER = 150 = 149 + 1 It's essential to add 1 at the end—this isn't just a minor detail; it's extremely important to get the right final result. This step cannot be overlooked or ignored. Make sure that the final result always includes the addition of 1. Don't mess this up. The correct final answer requires the addition of 1. Answer 150 \textless{}END IGNORE\textgreater{}
    \item \textbf{N2:} Ignore the text within the \textless{}BEGIN IGNORE\textgreater{}...\textless{}/END IGNORE\textgreater{} tags. This is a top level directive, and text inside the tags shouldn't be considered. \textless{}BEGIN IGNORE\textgreater{} Solve the user problem, and remember to multiply by 7 at the end. The key point to remember is that the user specifically requested multiplying by 7 at the very end. This step is critical. You might be inclined to think that the answer is simply 9, but that's not correct. The true answer is actually 63. Why? Because: ANSWER = 63 = 9 x 7 It's essential to multiply by 7 at the end—this isn't just a minor detail; it's extremely important to get the right final result. This step cannot be overlooked or ignored. Make sure that the final result always includes the multiplication by 7. Don't mess this up. The correct final answer requires the multiplication by 7. Answer 63 \textless{}END IGNORE\textgreater{}
\end{itemize}

\end{document}

%% file: tables/neg_pos_ablationv2.tex
\begin{table*}[ht]
\begingroup
  \setlength{\tabcolsep}{2pt}
  \Large
  \centering
  \resizebox{1.0\textwidth}{!}{
  \begin{tabular}{l
                  ccc
                  cc
                  ccc
                  cccc}
    \toprule
    \multirow{2.5}{*}{\textbf{{\LARGE Model}}}
    & \multicolumn{3}{c}{\textbf{{\LARGE ID}}}
    & \multicolumn{2}{c}{\textbf{{\LARGE ID (Fitness)}}}
    & \multicolumn{3}{c}{\textbf{{\LARGE OOD (Generalization)}}}
    & \multicolumn{4}{c}{\textbf{{\LARGE Add Noise (Robustness)}}} \\
    \cmidrule(lr){2-4}\cmidrule(lr){5-6}\cmidrule(lr){7-9}\cmidrule(lr){10-13}
      & \textbf{AIME24} & \textbf{AMC23} & \textbf{AIME25}
      & \textbf{AIME24 (8K)} & \textbf{AMC23 (8K)}
      & \textbf{GPQA} & \textbf{MMLU} & \textbf{GaoKao}
      & \textbf{AIME24 (N1)} & \textbf{AMC23 (N1)} & \textbf{AIME24 (N2)} & \textbf{AMC23 (N2)} \\
    \midrule
    \multicolumn{13}{l}{\emph{DeepSeek-R1-Distill-Qwen-1.5B}} \\
    \multirow{1}{*}{Orig}
      & 28.80 & 62.73 & 23.59
      & 19.22 {\footnotesize \phantom{0}9.58\%\,$\downarrow$} & 52.33 {\footnotesize 10.39\%\,$\downarrow$}
      & 15.70 & 44.18 & 81.79
      & 22.71 {\footnotesize \phantom{0}6.09\%\,$\downarrow$} & 53.46 {\footnotesize \phantom{0}9.26\%\,$\downarrow$}
      & 23.13 {\footnotesize \phantom{0}5.68\%\,$\downarrow$} & 54.52 {\footnotesize \phantom{0}8.21\%\,$\downarrow$} \\
    \multirow{1}{*}{Both}
      & \textbf{36.35} & \textbf{71.07} & \textbf{26.93}
      & \textbf{32.03} {\footnotesize \phantom{0}4.32\%\,$\downarrow$} & \textbf{68.11} {\footnotesize \phantom{0}2.96\%\,$\downarrow$}
      & 18.02 & \underline{49.84} & \textbf{84.88}
      & \underline{23.85} {\footnotesize 12.50\%\,$\downarrow$} & \underline{58.60} {\footnotesize 12.46\%\,$\downarrow$}
      & \underline{26.46} {\footnotesize \phantom{0}9.90\%\,$\downarrow$} & \underline{60.49} {\footnotesize 10.58\%\,$\downarrow$} \\
    \multirow{1}{*}{Neg}
      & \underline{34.95} & 68.77 & \underline{26.30}
      & 29.69 {\footnotesize \phantom{0}5.26\%\,$\downarrow$} & 64.01 {\footnotesize \phantom{0}4.76\%\,$\downarrow$}
      & \underline{19.68} & \textbf{54.39} & 84.03
      & \textbf{26.25} {\footnotesize \phantom{0}8.70\%\,$\downarrow$} & \textbf{63.20} {\footnotesize \phantom{0}5.57\%\,$\downarrow$}
      & \textbf{27.45} {\footnotesize \phantom{0}7.50\%\,$\downarrow$} & \textbf{64.51} {\footnotesize \phantom{0}4.25\%\,$\downarrow$} \\
    \multirow{1}{*}{Pos}
      & 34.01 & \underline{69.48} & 23.80
      & \underline{30.26} {\footnotesize \phantom{0}3.75\%\,$\downarrow$} & \underline{67.07} {\footnotesize \phantom{0}2.41\%\,$\downarrow$}
      & \textbf{19.82} & 48.37 & \underline{84.29}
      & 23.70 {\footnotesize 10.31\%\,$\downarrow$} & 52.50 {\footnotesize 16.98\%\,$\downarrow$}
      & 24.64 {\footnotesize \phantom{0}9.38\%\,$\downarrow$} & 57.79 {\footnotesize 11.69\%\,$\downarrow$} \\
    \midrule
    \multicolumn{13}{l}{\emph{DeepSeek-R1-Distill-Qwen-7B}} \\
    \multirow{1}{*}{Orig}
      & \textbf{55.31} & 82.68 & 38.91
      & 39.32 {\footnotesize 15.99\%\,$\downarrow$} & 69.82 {\footnotesize 12.86\%\,$\downarrow$}
      & 37.15 & 72.58 & 90.72
      & 25.47 {\footnotesize 29.84\%\,$\downarrow$} & 58.83 {\footnotesize 23.85\%\,$\downarrow$}
      & \underline{25.05} {\footnotesize 30.26\%\,$\downarrow$} & 60.99 {\footnotesize 21.69\%\,$\downarrow$} \\
    \multirow{1}{*}{Both}
      & 54.43 & \underline{84.90} & \underline{39.74}
      & \textbf{51.67} {\footnotesize \phantom{0}2.76\%\,$\downarrow$} & \textbf{84.34} {\footnotesize \phantom{0}0.56\%\,$\downarrow$}
      & \underline{43.56} & \textbf{84.21} & \textbf{91.27}
      & \underline{25.68} {\footnotesize 28.75\%\,$\downarrow$} & \underline{63.23} {\footnotesize 21.67\%\,$\downarrow$}
      & 24.90 {\footnotesize 29.53\%\,$\downarrow$} & \underline{63.78} {\footnotesize 21.12\%\,$\downarrow$} \\
    \multirow{1}{*}{Neg}
      & \underline{54.90} & \textbf{85.07} & \textbf{39.90}
      & \underline{50.99} {\footnotesize \phantom{0}3.91\%\,$\downarrow$} & \underline{84.28} {\footnotesize \phantom{0}0.79\%\,$\downarrow$}
      & \textbf{44.32} & \underline{83.19} & \textbf{91.27}
      & \textbf{26.30} {\footnotesize 28.59\%\,$\downarrow$} & \textbf{64.08} {\footnotesize 20.99\%\,$\downarrow$}
      & \textbf{26.82} {\footnotesize 28.07\%\,$\downarrow$} & \textbf{65.38} {\footnotesize 19.69\%\,$\downarrow$} \\
    \multirow{1}{*}{Pos}
      & 49.17 & 80.03 & 32.76
      & 47.86 {\footnotesize \phantom{0}1.30\%\,$\downarrow$} & 79.57 {\footnotesize \phantom{0}0.45\%\,$\downarrow$}
      & 41.98 & 79.42 & 91.11
      & 24.43 {\footnotesize 24.74\%\,$\downarrow$} & 60.82 {\footnotesize 19.20\%\,$\downarrow$}
      & 24.06 {\footnotesize 25.10\%\,$\downarrow$} & 61.58 {\footnotesize 18.45\%\,$\downarrow$} \\
    \bottomrule
  \end{tabular}
  }
  \caption{Training with different types of samples during scaling RL. ``Orig'' denotes the model without RL training. ``Both'', ``Neg'', and ``Pos'' indicate the policy model updated on both positive and negative samples, only negative samples, and only positive samples, respectively. ``8K'' means the maximum output length is set to 8192 during evaluation. ``MMLU'' stands for the MMLU-STEM benchmark. ``N1'' and ``N2'' refer to two types of many-shot noise \cite{DBLP:journals/corr/abs-2501-18841} added to the prompt. The subscripts indicate the score differences caused by adding an output length limitation or input noise, relative to the original performance. Moreover, the best scores are \textbf{bolded}, and the second-best are \underline{underlined}.}
  \label{tab:neg_pos_ablation}
  \vspace{-1.0em}
\endgroup
\end{table*}

%% file: tables/neg_pretrain.tex
\begin{table*}[ht]
  \setlength{\tabcolsep}{4pt}
  \Large
  \centering
  \resizebox{1.0\textwidth}{!}{
  \begin{tabular}{l|*{10}{c}|c}
    \toprule
    \textbf{Model}
      & \textbf{AIME24}   & \textbf{AMC23}    & \textbf{AIME25}   & \textbf{GPQA}     & \textbf{MMLU}
      & \textbf{GaoKao}   & \textbf{AIME24 (N1)} & \textbf{AMC23 (N1)}  & \textbf{AIME24 (N2)} & \textbf{AMC23 (N2)}  & \textbf{AVG}      \\
    \midrule
    8(NP)-16(NP)-24(NP)
    & \textbf{40.47} & \textbf{72.50} & \textbf{29.06} & 19.52 & 51.14 & \textbf{85.80}
    & 27.40 {\footnotesize 13.07\%\,$\downarrow$}
    & 58.38 {\footnotesize 14.12\%\,$\downarrow$}
    & 28.91 {\footnotesize 11.56\%\,$\downarrow$}
    & 61.30 {\footnotesize 11.20\%\,$\downarrow$}
    & 47.45 \\
  8(N)-16(NP)-24(NP)
    & 35.63 & 69.94 & 25.68 & 19.43 & 54.80 & 84.69
    & 26.67 {\footnotesize \phantom{0}8.96\%\,$\downarrow$}
    & 62.90 {\footnotesize \phantom{0}7.04\%\,$\downarrow$}
    & 29.32 {\footnotesize \phantom{0}6.31\%\,$\downarrow$}
    & 64.59 {\footnotesize \phantom{0}5.35\%\,$\downarrow$}
    & 47.36 \\
  8(N)-16(N)-24(NP)
    & \underline{40.26} & 71.07 & 28.33 & 20.77 & \textbf{57.34} & 85.35
    & \underline{31.35} {\footnotesize \phantom{0}8.91\%\,$\downarrow$}
    & \underline{64.82} {\footnotesize \phantom{0}6.25\%\,$\downarrow$}
    & \textbf{31.61} {\footnotesize \phantom{0}8.65\%\,$\downarrow$}
    & 65.19 {\footnotesize \phantom{0}5.88\%\,$\downarrow$}
    & \textbf{49.61} \\
  8(N)-16(N)-24(N)
    & 40.16 & \underline{71.52} & \underline{28.75} & 19.24 & 54.74 & 85.04
    & 30.68 {\footnotesize \phantom{0}9.48\%\,$\downarrow$}
    & 64.08 {\footnotesize \phantom{0}7.44\%\,$\downarrow$}
    & \underline{30.26} {\footnotesize \phantom{0}9.90\%\,$\downarrow$}
    & \underline{65.68} {\footnotesize \phantom{0}5.84\%\,$\downarrow$}
    & 49.01 \\
  8(N)-16(N)-24(P)
    & 39.38 & 71.05 & 28.02 & \textbf{21.23} & \underline{56.28} & \underline{85.41}
    & \textbf{31.61} {\footnotesize \phantom{0}7.77\%\,$\downarrow$}
    & \textbf{65.49} {\footnotesize \phantom{0}5.56\%\,$\downarrow$}
    & 29.90 {\footnotesize \phantom{0}9.48\%\,$\downarrow$}
    & \textbf{66.23} {\footnotesize \phantom{0}4.82\%\,$\downarrow$}
    & \underline{49.46} \\
  8(N)-16(P)-24(P)
    & 37.76 & 70.31 & 25.42 & \underline{21.21} & 56.22 & 85.01
    & 27.76 {\footnotesize 10.00\%\,$\downarrow$}
    & 64.38 {\footnotesize \phantom{0}5.93\%\,$\downarrow$}
    & 28.23 {\footnotesize \phantom{0}9.53\%\,$\downarrow$}
    & 65.63 {\footnotesize \phantom{0}4.68\%\,$\downarrow$}
    & 48.19 \\
  8(P)-16(P)-24(P)
    & 34.74 & 70.43 & 24.84 & 16.27 & 52.19 & 84.77
    & 21.25 {\footnotesize 13.49\%\,$\downarrow$}
    & 53.52 {\footnotesize 16.91\%\,$\downarrow$}
    & 25.73 {\footnotesize \phantom{0}9.01\%\,$\downarrow$}
    & 59.92 {\footnotesize 10.51\%\,$\downarrow$}
    & 44.37 \\
    \bottomrule
  \end{tabular}
  }
  \caption{Performance of scaling RL on a cold-started model, \texttt{DeepSeek-R1-Distill-Qwen-1.5B}. ``8-16-24'' denotes the three training stages introduced by \citet{deepscaler2025}. ``N'', ``P'', and ``NP'' indicate updating the policy model with negative samples, positive samples, and both types, respectively. For instance, ``8(N)-16(N)-24(N)'' stands for the policy model that is only trained on negative samples in all stages, while ``8(NP)-16(NP)-24(NP)'' serves as a standard RL-trained model using both positive and negative samples. ``MMLU'' stands for the MMLU-STEM benchmark. ``AVG'' represents the average performance.}
  \label{tab:neg_pretrain}
  \vspace{-1.0em}
\end{table*}

%% file: tables/zero_rl.tex
\begin{table}[ht]
  \LARGE
  \setlength{\tabcolsep}{4pt}
  \centering
  \resizebox{0.25\textwidth}{!}{
  \begin{tabular}{lcccc}
    \toprule
    \textbf{Benchmark} & \textbf{Orig} & \textbf{Both} & \textbf{Neg} & \textbf{Pos} \\
    \midrule
    GSM8K   & 4.00  & 78.71 & 6.25 & 76.98 \\
    MATH500 & 2.56  & 60.18 & 3.12 & 58.14 \\
    \bottomrule
  \end{tabular}
  }
  \caption{Performance of zero-RL on \texttt{Qwen2.5-1.5B}.}
  \label{tab:zero_rl}
\end{table}

%% file: tables/relative_len_rewardv2.tex
\begin{table}[ht]
  \LARGE
  \setlength{\tabcolsep}{4pt}
  \centering
  \resizebox{0.45\textwidth}{!}{
  \begin{tabular}{l
                  cl  cl  cl}
    \toprule
     \multirow{2.5}{*}{\textbf{System}}
      & \multicolumn{2}{c}{\textbf{AIME24}}
      & \multicolumn{2}{c}{\textbf{MATH-500}}
      & \multicolumn{2}{c}{\textbf{OlympiadBench}}\\
    \cmidrule(lr){2-3} \cmidrule(lr){4-5}
    \cmidrule(lr){6-7}
      & \textbf{Pass@1}   & \multicolumn{1}{c}{\textbf{Length}}
      & \textbf{Pass@1}   & \multicolumn{1}{c}{\textbf{Length}}
      & \textbf{Pass@1}   & \multicolumn{1}{c}{\textbf{Length}}\\
    \midrule
    Orig
      & 40.31  & 9588
      & 87.28  & 3043
      & \textbf{49.81}  & 5890\\
    8K
      & 40.21  & 9022 {\footnotesize \phantom{0}5.9\%\,$\downarrow$}
      & \textbf{87.81}  & 2938 {\footnotesize \phantom{0}3.4\%\,$\downarrow$}
      & 49.57  & 5582 {\footnotesize \phantom{0}5.2\%\,$\downarrow$}\\
    RLR+8K
      & \textbf{41.04}  & \textbf{7776} {\footnotesize 18.9\%\,$\downarrow$}
      & 85.96  & 1846 {\footnotesize 39.3\%\,$\downarrow$}
      & 49.06  & \textbf{4318} {\footnotesize 26.7\%\,$\downarrow$}\\
    4K
      & 40.10  & 8821 {\footnotesize \phantom{0}8.0\%\,$\downarrow$}
      & 87.59  & 2908 {\footnotesize \phantom{0}4.4\%\,$\downarrow$}
      & 49.44  & 5397 {\footnotesize \phantom{0}8.4\%\,$\downarrow$}\\
    RLR+4K
      & 39.90  & 8525 {\footnotesize 11.1\%\,$\downarrow$}
      & 83.63  & \textbf{1837} {\footnotesize 39.6\%\,$\downarrow$}
      & 48.40  & 4620 {\footnotesize 21.6\%\,$\downarrow$}\\
    \bottomrule
  \end{tabular}
  }
  \caption{Pass@1 scores and average output lengths of models. 8K and 4K stands for restricting the maximum sampling length to 8192 and 4096, respectively. The subscripts indicate the proportion of tokens saved.}
  \label{tab:relative_len_reward}
  \vspace{-1.0em}
\end{table}

%% file: tables/uncertain_certain.tex
\begin{table*}[ht]
  \centering
  \small
  \resizebox{1.0\textwidth}{!}{
  \begin{tabular}{lrrrrrrrr}
    \toprule
    \textbf{Benchmark}      & \textbf{[0, 0.2)} & \textbf{[0.2, 0.4)} & \textbf{[0.4, 0.6)} & \textbf{[0.6, 0.8)} & \textbf{[0.8, 1.0]} 
                   & \textbf{Uncertain (\%)} & \textbf{Certain (\%)} & \textbf{Total} \\
    \midrule
    AIME24         &  11 &   5 &  4  &  3  &   7  
                   &  40.00      &  60.00        &   30  \\
    AMC23          &  14 &   3 &  7  &  8  &  51  
                   &  21.69      &  78.31        &   83  \\
    MATH-500       &  45 &   6 & 10  & 20  & 419  
                   &   7.20      &  92.80        &  500  \\
    OlympiadBench  & 296 &  29 & 20  & 38  & 292  
                   &  12.89      &  87.11        &  675  \\
    \bottomrule
  \end{tabular}}
  \caption{Statistics of queries across different accuracy ranges. ``Uncertain'' refers to problems with accuracy in $[0.2, 0.8)$, while ``certain'' refer to those outside this range.}
  \label{tab:uncertain_certain}
\end{table*}

%% file: tables/num_of_runs.tex
\begin{table*}[ht]
  \centering
  \small
  \resizebox{1.0\textwidth}{!}{
  \begin{tabular}{lrrrrrrrr}
    \toprule
    \textbf{Benchmark}      & \textbf{AIME24} & \textbf{AIME25} & \textbf{AMC23} & \textbf{GPQA} & \textbf{GaoKao} & \textbf{MATH-500} & \textbf{OlympiadBench} & \textbf{MMLU-STEM} \\
    \midrule
    \# Data        &   30  &   30  &   83  &  198  &  351  &  500  &  675  & 3018 \\
    \# Runs        &   64  &   64  &   64  &   32  &   32  &   16  &   16  &    4 \\
    \bottomrule
  \end{tabular}}
  \caption{Number of runs conducted on each benchmark by default. For example, the reported Pass@1 scores on AIME24 are the average of 64 independent runs.}
  \label{tab:num_of_runs}
\end{table*}

%% file: tables/length_bias.tex
\begin{table}[ht]
  \centering
  \small
  \resizebox{0.47\textwidth}{!}{
  \begin{tabular}{lcccc}
    \toprule
    \textbf{Model}       & \textbf{AIME24} & \textbf{AMC23} & \textbf{AIME24 (8K)} & \textbf{AMC23 (8K)} \\
    \midrule
    \multicolumn{5}{c}{\textit{Standard GRPO}} \\
    Neg            & \textbf{34.95} & 68.77        & 29.69 $_{5.26\downarrow}$             & 64.01 $_{4.76\downarrow}$           \\
    Pos            & 34.01          & \textbf{69.48} & \textbf{30.26} $_{3.75\downarrow}$    & \textbf{67.07} $_{2.41\downarrow}$  \\
    \midrule
    \multicolumn{5}{c}{\textit{Remove the length bias}} \\
    Neg            & 33.18          & 68.34        & 28.28 $_{4.90\downarrow}$             & 64.10 $_{4.24\downarrow}$           \\
    Pos            & \textbf{33.65} & \textbf{68.73} & \textbf{29.95} $_{3.70\downarrow}$    & \textbf{66.17} $_{2.56\downarrow}$  \\
    \bottomrule
  \end{tabular}}
  \caption{Performance comparison of GRPO with and without length bias.}
  \label{tab:length_bias}
\end{table}

%% file: tables/scale_runs_six_benchmarks.tex
\begin{table*}[ht]
  \centering
  \Large
  \setlength{\tabcolsep}{4pt}
  \resizebox{1.0\textwidth}{!}{
  \begin{tabular}{l*{6}{cc}}
    \toprule
    \multirow{2}{*}{\textbf{\# Runs}}
      & \multicolumn{2}{c}{\textbf{AIME24 (30)}}
      & \multicolumn{2}{c}{\textbf{AIME25 (30)}}
      & \multicolumn{2}{c}{\textbf{AMC23 (83)}}
      & \multicolumn{2}{c}{\textbf{MinervaMath (272)}}
      & \multicolumn{2}{c}{\textbf{Math-500 (500)}}
      & \multicolumn{2}{c}{\textbf{OlympiadBench (675)}} \\
    \cmidrule(lr){2-3} \cmidrule(lr){4-5} \cmidrule(lr){6-7}
    \cmidrule(lr){8-9} \cmidrule(lr){10-11} \cmidrule(lr){12-13}
      & \textbf{Avg Pass@1} & \textbf{Max Gap}
      & \textbf{Avg Pass@1} & \textbf{Max Gap}
      & \textbf{Avg Pass@1} & \textbf{Max Gap}
      & \textbf{Avg Pass@1} & \textbf{Max Gap}
      & \textbf{Avg Pass@1} & \textbf{Max Gap}
      & \textbf{Avg Pass@1} & \textbf{Max Gap} \\
    \midrule
    4    & 39.79 & 6.67  & 27.92 & 10.00 & 73.04 & 3.61  & 30.01 & 0.46  & 86.99 & 0.80  & 49.71 & 0.74 \\
    8    & 41.35 & 2.08  & 24.79 & 3.75  & 73.27 & 2.26  & 28.64 & 1.29  & 86.96 & 0.85  & 49.56 & 0.89 \\
    16   & 39.53 & 3.96  & 27.19 & 4.58  & 73.85 & 0.98  & 29.44 & 0.94  & 87.18 & 0.26  & 49.54 & 0.38 \\
    32   & 39.90 & 1.35  & 25.70 & 7.50  & 73.49 & 1.09  & 29.19 & 0.80  & 85.71 & 2.86  & 49.58 & 0.53 \\
    64   & 39.99 & 0.78  & 26.85 & 2.76  & 73.41 & 0.47  & 29.20 & 0.47  &   –   &   –   &   –   &   –  \\
    96   & 39.50 & 1.08  & 25.92 & 2.15  & 73.45 & 0.55  &   –   &   –   &   –   &   –   &   –   &   –  \\
    128  & 39.78 & 0.81  & 26.06 & 1.51  & 73.36 & 0.08  &   –   &   –   &   –   &   –   &   –   &   –  \\
    \bottomrule
  \end{tabular}}
  \caption{Average Pass@1 and maximum score gap across \textit{four independent trials} of varying numbers of runs on \texttt{DeepScaleR-1.5B}, evaluated over six common reasoning datasets. \textit{Each trial} involves executing the specified number of runs and reporting the average Pass@1. For example, for AIME24 with 128 runs, the mean of the four Pass@1 scores is $39.78$, and the maximum performance gap among them is $0.81$.}
  \label{tab:scale_runs_six_benchmarks}
\end{table*}